\documentclass[smallabstract,smallcaptions]{dccpaper}

\usepackage{epsfig}
\usepackage{citesort}
\usepackage{amsmath}
\usepackage{amssymb}
\usepackage{color}
\usepackage{url}
\usepackage{epstopdf}
\usepackage{multirow}
\usepackage{booktabs}
\usepackage{makecell}
\usepackage{algorithm}  
\usepackage{algorithmicx} 
\usepackage{algpseudocode}  
\usepackage{amsmath}  
 
\newlength{\figurewidth}
\newlength{\smallfigurewidth}
\usepackage{subfigure}

\begin{document}

\title
{\large
\textbf{Optimized Adaptive Loop Filter in Versatile Video Coding}
}

\author{%
Xuewei Meng$^{\ast}$, Jiaqi Zhang$^{\dag}$, Chuanmin Jia$^{\ast}$, Xinfeng Zhang$^{\dag}$,\\
Shanshe Wang$^{\ast}$ and Siwei Ma$^{\ast}$\\[0.5em]
{\small\begin{minipage}{\linewidth}\begin{center}
\begin{tabular}{ccc}
$^{\ast}$Institute of Digital Media & \hspace*{0.1in} & $^{\dag}$Key Laboratory of Intelligent Information Processing \\
Peking University &&  Institute of Computing Technology  \\
Beijing, China && Beijing, China\\
\url{xwmeng@pku.edu.cn} && \url{zhangjiaqi17@mails.ucas.ac.cn}
\end{tabular}
\end{center}\end{minipage}}
}

\maketitle
\thispagestyle{empty}

\begin{abstract}
In the Versatile Video Coding~(VVC) standard, adaptive loop filter~(ALF), including Geometry transformation-based Adaptive Loop Filter~(GALF) and Cross Component Adaptive Loop Filter~(CCALF), plays an essential role in reducing compression artifacts. However, it also has high coding complexity and requires many picture buffer accesses in the encoder that will increase external memory access and is unfriendly to the software and hardware design. Therefore, we propose an optimized ALF framework, including the parallel design of GALF and CCALF, the adaptive parameter decision of GALF, and one-pass CCALF scheme by effectively estimating the CCALF filtering distortion without conducting filter operation. Compared to VTM-8.0, the proposed method can reduce the picture buffer access from 152 to 1 and achieve roughly 25\% time-savings of the ALF module with negligible coding performance change under RA configuration. Some of the proposed methods have been adopted in the VVC reference software.

\end{abstract}

\makeatletter
\newcommand{\thickhline}{%
	\noalign {\ifnum 0=`}\fi \hrule height 1pt
	\futurelet \reserved@a \@xhline
}

\Section{Introduction}
In-loop filter has been recognized as an essential module in the hybrid video coding framework because of its capability in reducing compression artifacts, such as blocking, ringing, and blurring artifacts. There are two kinds of in-loop filters in  H.265/HEVC~\cite{HEVC}, i.e., Deblocking Filter~\cite{DF} and Sample Adaptive Offset~(SAO)~\cite{SAO}. To further improve the subjective and objective quality of the compressed video, adaptive loop filter~(ALF) module, including Geometry transformation-based Adaptive Loop Filter~(GALF)~\cite{GALF,NonlinearALF-1,NonlinearALF-2} and Cross Component Adaptive Loop Filter~(CCALF)~\cite{CC-ALF,CCALF-1,CCALF-2,CCALF22,r0322,R0225}, is adopted in Versatile Video Coding~(VVC), a new video coding standard, developed by Joint Video Experts Team~(JVET)~\cite{CFP} since October 2017. In addition, non-local-based in-loop filters~\cite{non-local1,non-local2,non-local3,non-local4} are investigated during the standardization process.

ALF, a Wiener-based filter, trains filter coefficients in the encoder by using the reconstructed samples after SAO and the original samples according to the principle of minimizing the mean square error~(MSE). Then the filter coefficients derived in the encoder are transmitted to the decoder for the ALF decoding process. During the development of HEVC, several ALFs were investigated, such as block-based ALF~(BALF)~\cite{ALF-inloop} and directional ALF~(DALF)~\cite{zheng2011directional}. To derive a low-complexity and high-adaptability adaptive loop filter, GALF was proposed in \cite{GALF}. This method introduces geometric transformations, such as rotation, diagonal and vertical flip, to the samples depending on the orientation of the gradient of the reconstructed samples after SAO. With the introduction of geometric transformations, more spatial adaptation is supported without excessive signaling of filter coefficients. The geometric transformation scheme is adopted in the ALF of VVC. 

The ALFs mentioned above, including BALF, DALF, and GALF, are all designed for the luma component. While for chroma components, only one filter is used for a Coding Tree Unit~(CTU), limiting the performance of chroma ALF. To improve the performance of chroma components without improving the complexity, a CCALF techonology~\cite{CC-ALF} was adopted in VVC to improve the chroma fidelity by exploiting correlations between the luma and chroma channels. CCALF applies a linear filter to the luma channel to determine a chroma residual that is added to a previously reconstructed chroma signal.

The current ALF module in VVC includes GALF and CCALF. As the luma sample used by CCALF is before GALF, CCALF and GALF can be parallel in the decoder. However, they can not be conducted in parallel in the encoder. Typically, although ALF takes up no more than 10\%~\cite{R0013} of the encoding time in VTM-8.0~\cite{VTM8.0}, its complex and parallel-unfriendly parameter training process makes it unpractical in the fast preset of real-time encoders, such as the open optimized VVC encoder~\cite{fraunhoferhhi}. The dependency between GALF and CCALF makes it more difficult for the real-time encoder design. Moreover, CCALF is a multi-pass process in the encoder. One encoding pass requires an access of picture buffer. CCALF needs up to 152 picture buffer accesses. Picture buffers are usually off-chip and allocated in external memory because of the limited on-chip memory. The multi-pass CCALF indicates massive external memory bandwidth, and one off-chip data access may require 10 times of power consumption and latency compared to one on-chip data access~\cite{tsai2013adaptive}. Therefore, the multi-pass CCALF encoding algorithm should be avoided.

%Specifically, the selection of best CCALF coefficients and the decision of CTU level filter index and on/off control are based on a Rate-Distortion Optimization~(RDO) method. The Rate-Distortion Cost~(RDCost) is calculated by,

%\begin{equation}\label{RDO}
%J_{RD} = D + \lambda \cdot R,
%\end{equation}
%where $R$ represents the rate introduced by the filter coefficients and on/off flags. $D$ is the distortion between the original image and the reconstructed image, which is derived by the CCALF filtering process. 

To solve the problems mentioned above, we propose a novel ALF framework, including the parallel design of GALF and CCALF, the adaptive parameter decision of GALF and one-pass CCALF scheme. Some of the proposed methods have been adopted in the reference software of VVC.

\Section{Background and Analysis}
In this section, we briefly review the encoder flow of ALF module first. Then encoding process of GALF is elaborated and analyzed to direct our optimization scheme. Finally, we detail the multi-pass encoding scheme of CCALF.

\begin{figure}[t!]
	\begin{center}
		\noindent
		\includegraphics[width=4.0in]{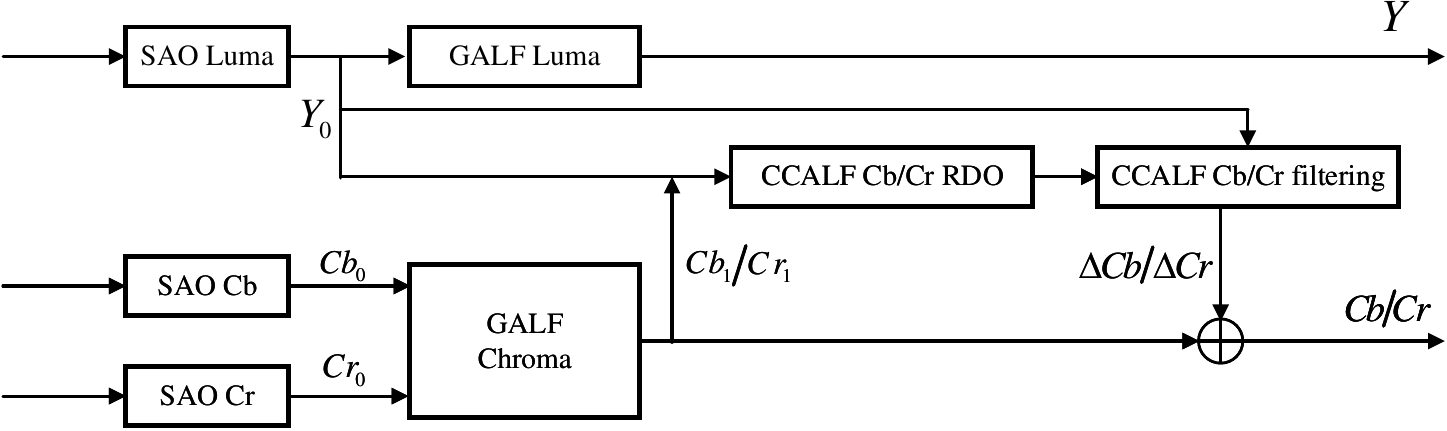}
        \vspace{-4mm}
		\caption{Framework of the ALF module in VTM-8.0.}\label{framework_enc}
	\end{center}
	\vspace{-12mm}
\end{figure}

\SubSection{2.1 Introduction of the Adaptive Loop Filter Module}	
The ALF in VVC comprises of GALF and CCALF, as shown in Fig.~\ref{framework_enc}. The GALF and CCALF filtering process both utilize the reconstructed frame after SAO, so that the two coding tools can be executed in parallel in the decoder. However, in the encoder, reconstructed chroma signal after ALF is needed in the CCALF filter coefficients training process. Hence, the two coding tools can not be performed in parallel.

%\begin{figure}[t!]
%	\begin{center}
%		\noindent
%		\includegraphics[width=4.5in]{Figures/framework_dec}
%		\caption{Framework of ALF module in the decoder of VTM-8.0.}\label{framework_dec}
%	\end{center}
%	\vspace{-8mm}
%\end{figure}

In the GALF, the filtering process is independent of each component~(Y, U and V). For the luma component, up to 25 filters are derived and one filter is selected for each $4\times4$ block based on the local gradients. For the chroma component, only one filter is enabled for a CTU.
% For chroma component, only one filter is used for a LCU. Two diamond filter shapes, as shown in Fig.~\ref{NALF-shape}, are used in GALF. The $7\times7$ diamond shape is applied for the luma component and the $5\times5$ diamond shape is used for chroma components.

%\begin{figure}[ht!]
%	\begin{center}
%		\noindent
%		\subfigure[5x5 shape for chroma]{
%			\includegraphics[width=1.32in]{Figures/5x5}
%		}
%		\subfigure[7x7 shape for luma]{
%			\includegraphics[width=1.88in]{Figures/7x7}
%		}
%		\caption{Filter shape of GALF.}\label{NALF-shape}
%	\end{center}
%\end{figure}

%As shown in Fig.~\ref{CCALF-shape}, a $4\times 3$ filter shape is used for the CCALF luma component, where $C0 \sim C7$ represent the filter coefficients.

%\begin{figure}[ht!]
%	\begin{center}
%		\noindent
%		\includegraphics[width=3.0in]{Figures/CCALF-shape}
%		\caption{Filter shape of CCALF.}\label{CCALF-shape}
%	\end{center}
%	\vspace{-8mm}
%\end{figure}

%\begin{figure}[ht!]
%	\begin{center}
%		\noindent
%		\includegraphics[width=4.0in]{Figures/CCALF-shape2}
%		\caption{Relative location of chroma sample being filtered and 
%			its reference luma samples for 4:2:0 format.}\label{relative-CCALF}
%	\end{center}
%	\vspace{-8mm}
%\end{figure}
Unlike the independent component processing of GALF, CCALF uses luma samples to refine each chroma component by applying a linear filter to the luma channel and then using the output of this filtering process for chroma components.  The main purpose of CCALF is to minimize the MSE between the to-be-filtered chroma samples and the original chroma samples using the reference luma samples. Filtered chroma sample $f[\textbf{k}]$ can be derived by
\begin{equation}
f[\textbf{k}] = c[\textbf{k}] + \sum_{n=0}^{N-1} w_nl[\textbf{k}^{'}+\textbf{p}_n].
\end{equation}
where $\textbf{k} = (x, y)$ denotes the sample location belonging to the to-be-filtered region $K$. $\textbf{k}^{'} = (x^{'}, y^{'})$ represents the location of the collocated luma sample. $l[\textbf{k}^{'}]$ is the collocated luma sample~(before GALF) of $c[\textbf{k}]$, the to-be filtered chroma sample~(after GALF). The filter coeffcients are described by $\textbf{w} = [w_0 \ w_1 \ \dots \ w_{N-1}]^{T}$. And  $\textbf{p}_n$ denotes the sample location offset to $\textbf{k}^{'}$ of the $n$ filter tap, $N$ is the number of filter tap.

%In order to find the minimum MSE between $s[\textbf{k}]$ and $f[\textbf{k}]$, we can calculate the derivatives of MSE with respect to $w_n$ and let the derivatives equal to zero. Then the Wiener-Hopf equations in a matrix form can be derived, as shown in Equ.~(\ref{Wiener}). In (\ref{Wiener}), $\textbf{R}_{l,l}$, $\textbf{w}$ and $\textbf{R}_{l,s-c}$ denote the auto-correlation matrix of the collocated luma samples $l[\textbf{k}^{'}]$, the filter coefficients, and the cross-correlation vector of the collocated luma samples $l[\textbf{k}^{'}]$ and the difference between original chroma samples $s[\textbf{k}]$ and to-be-filtered chroma samples $c[\textbf{k}]$, respectively. 

%\begin{equation}\label{Wiener}
% \textbf{w} = \textbf{R}_{l,l}^{-1} \cdot \textbf{R}_{l,s-c}.
%\end{equation}

\begin{figure}[t!]
	\begin{center}
		\noindent
		\includegraphics[width=5.0in]{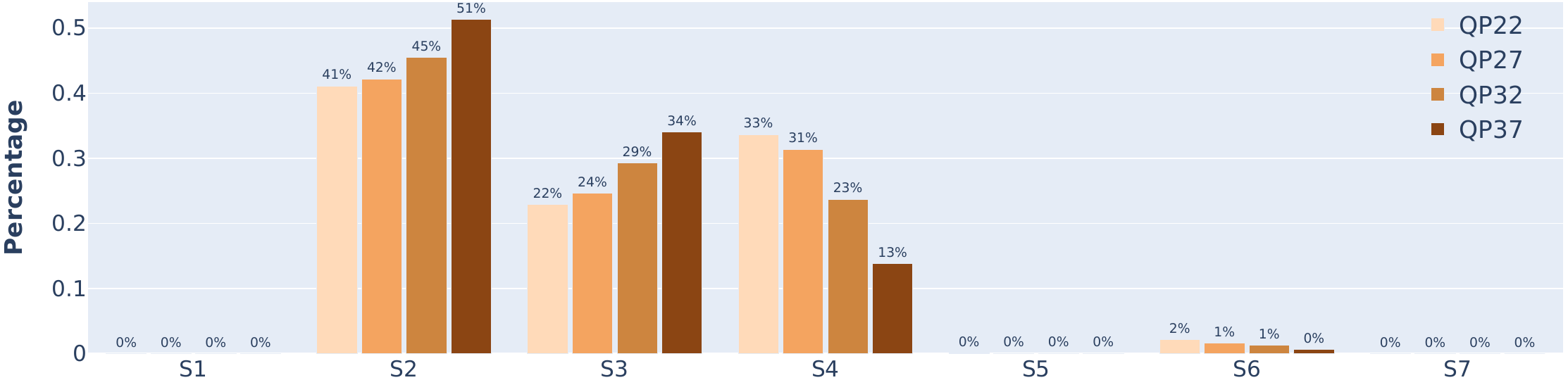}
        \vspace{-4mm}
		\caption{Percentage of the encoding time for each step in the ALF module.}\label{GALF-Encoding}
	\end{center}
	\vspace{-4mm}
\end{figure}

\begin{figure}[t!]
	\begin{center}
		\noindent
		\begin{tabular}{cc}
			\epsfig{width=2in,file=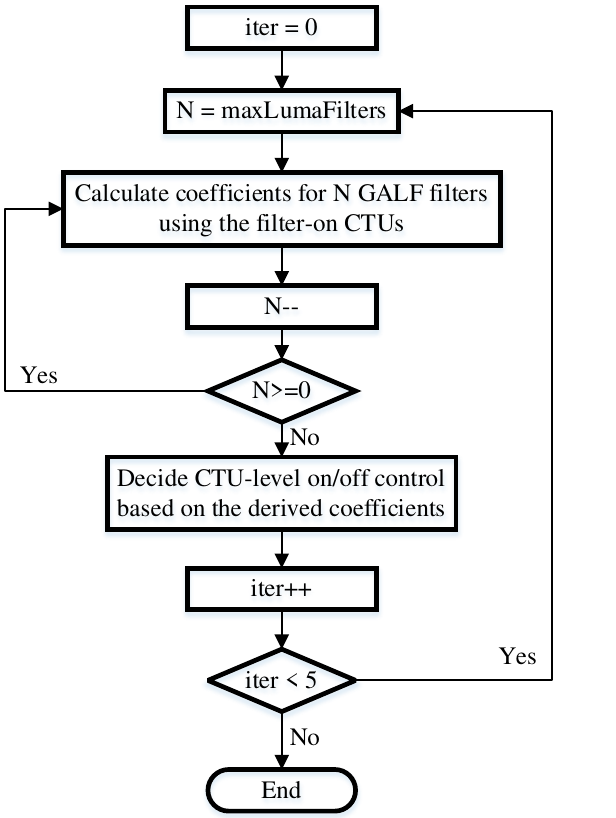} & 
			\epsfig{width=2in,file=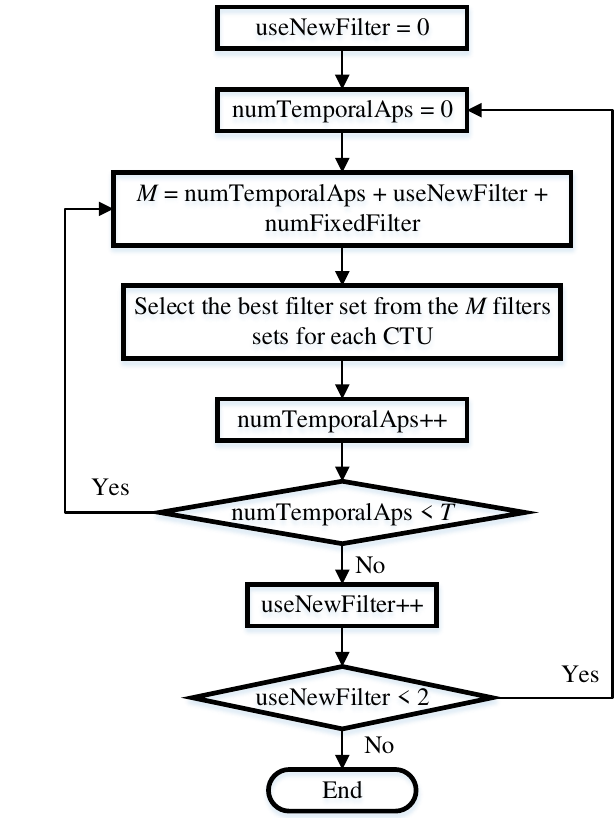} \\
			{\small (a)~GALF coefficient parameter training} & {\small (b)~GALF CTU level filter selection}
		\end{tabular}
	\end{center}
    \vspace{-4mm}
	\caption{Flowchart of GALF coefficient parameter training~(\textit{S3}) and CTU level filter selection~(\textit{S4}).} \label{Step3and4}
	\vspace{-5mm}
\end{figure}

\SubSection{2.2 Analysis of GALF Encoding Scheme}
In the encoding process of ALF, there are mainly seven steps, including $4\times4$ block-based classification, the calculation of auto-correlation matrix and cross-correlation vector, filter coefficient training, CTU-level filter selection, GALF filtering process with the selected parameters, CCALF coefficient training and CTU-level decision, and CCALF filtering process with the selected parameters. The first 5 steps are used for the GALF process labeled by \textit{S1} to \textit{S5}. Moreover, the remaining steps are for CCALF labeled by \textit{S6} and \textit{S7}. To get the most time-consuming steps, we statistic the time distribution of each step in VTM-8.0 under RA configuration. The test conditions follow the Common Test Condition~(CTC)~\cite{CTC}. As shown in Fig.~\ref{GALF-Encoding}, it is clear that the \textit{S2}, \textit{S3} and \textit{S4} are the most time-consuming steps. As \textit{S2} can be conducted in parallel for each CTU row, the complexity of \textit{S2} is usually lower than \textit{S3} and \textit{S4} in the real-time encoder. Hence, \textit{S3} and \textit{S4} are the bottlenecks of the ALF encoding process and the optimization of GALF lies in the complexity reduction of these two most complicated steps. The flowchart of these two steps is shown in Fig.~\ref{Step3and4}.

\begin{figure}[t!]
	\begin{center}
		\noindent
		\begin{tabular}{cccc}
			\epsfig{width=1.35in,file=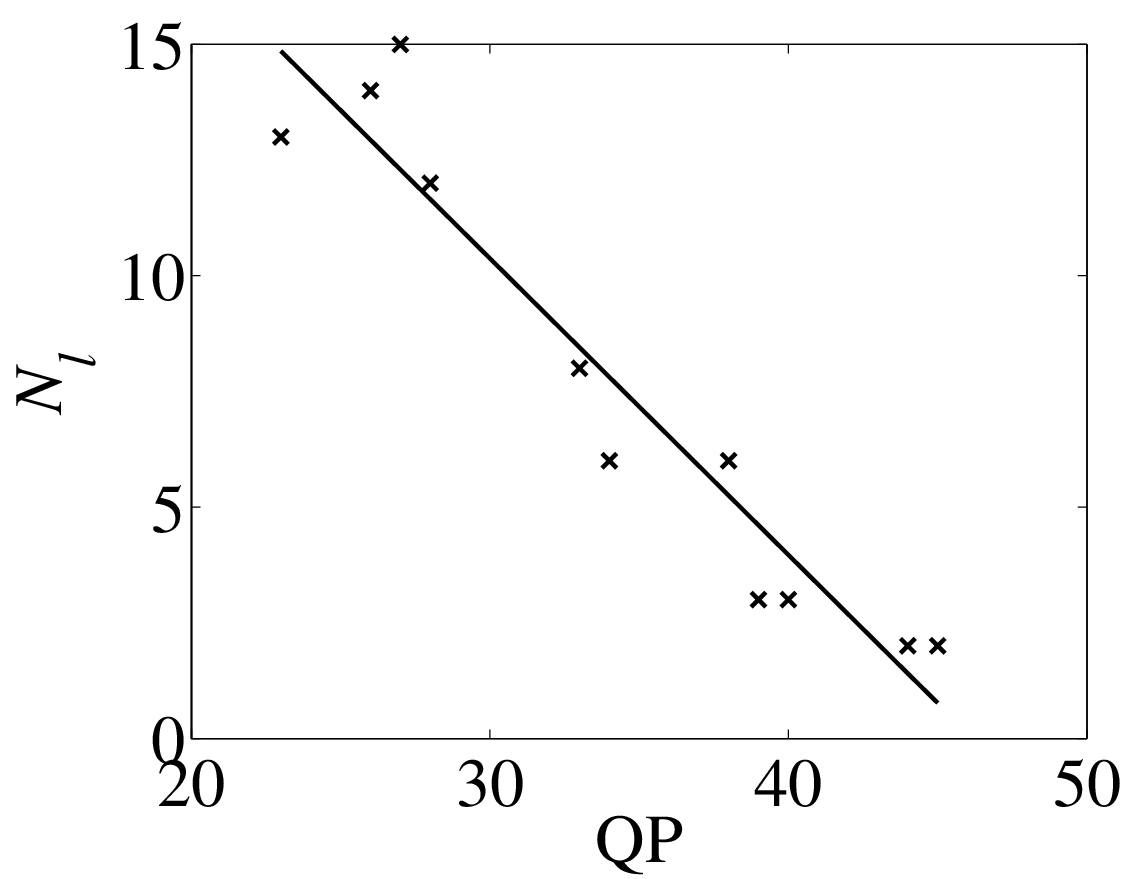} & 
			\epsfig{width=1.35in,file=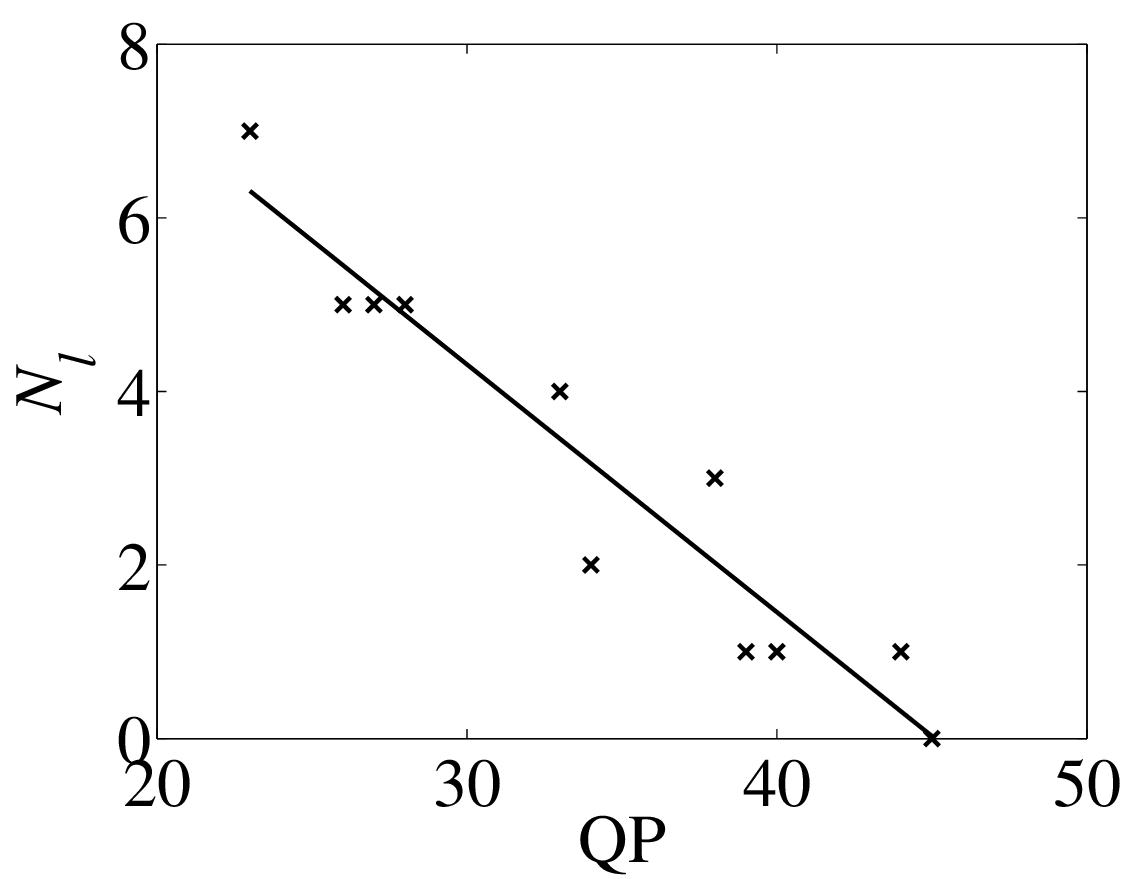} &
			\epsfig{width=1.35in,file=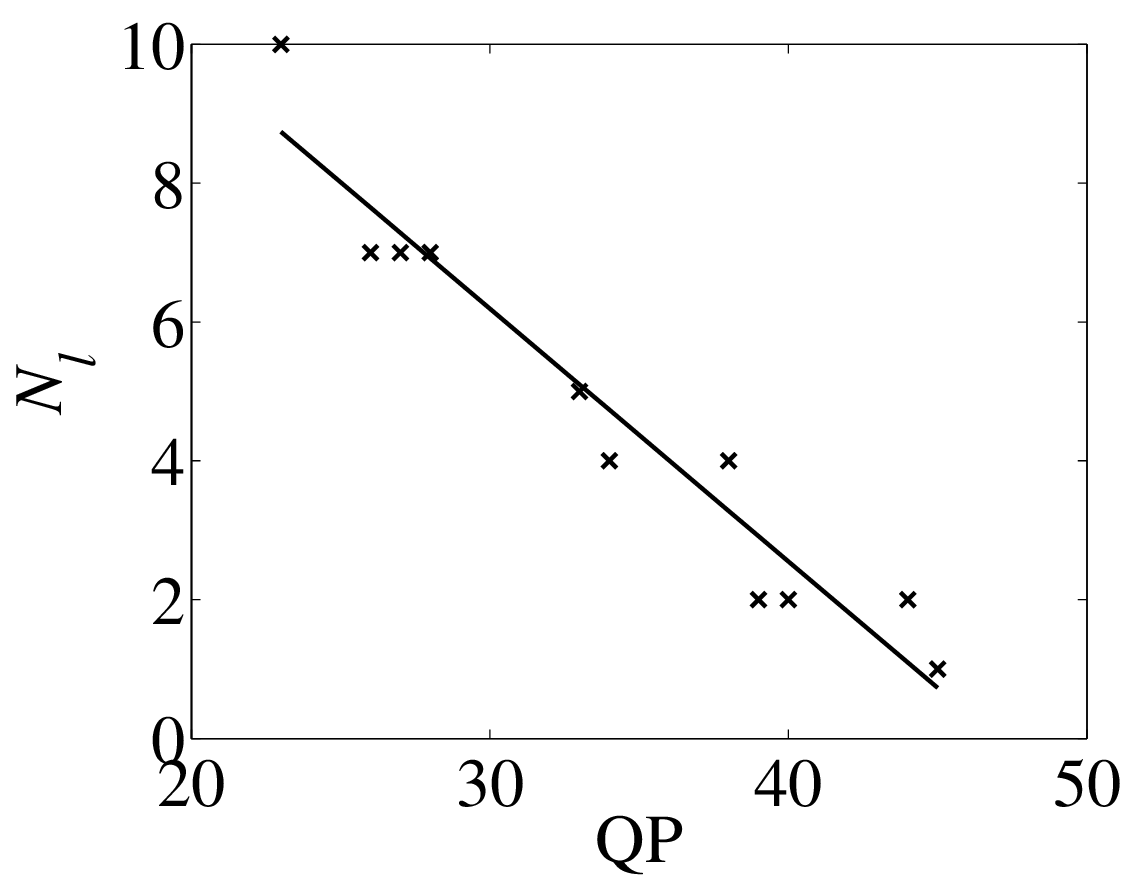} &
			\epsfig{width=1.35in,file=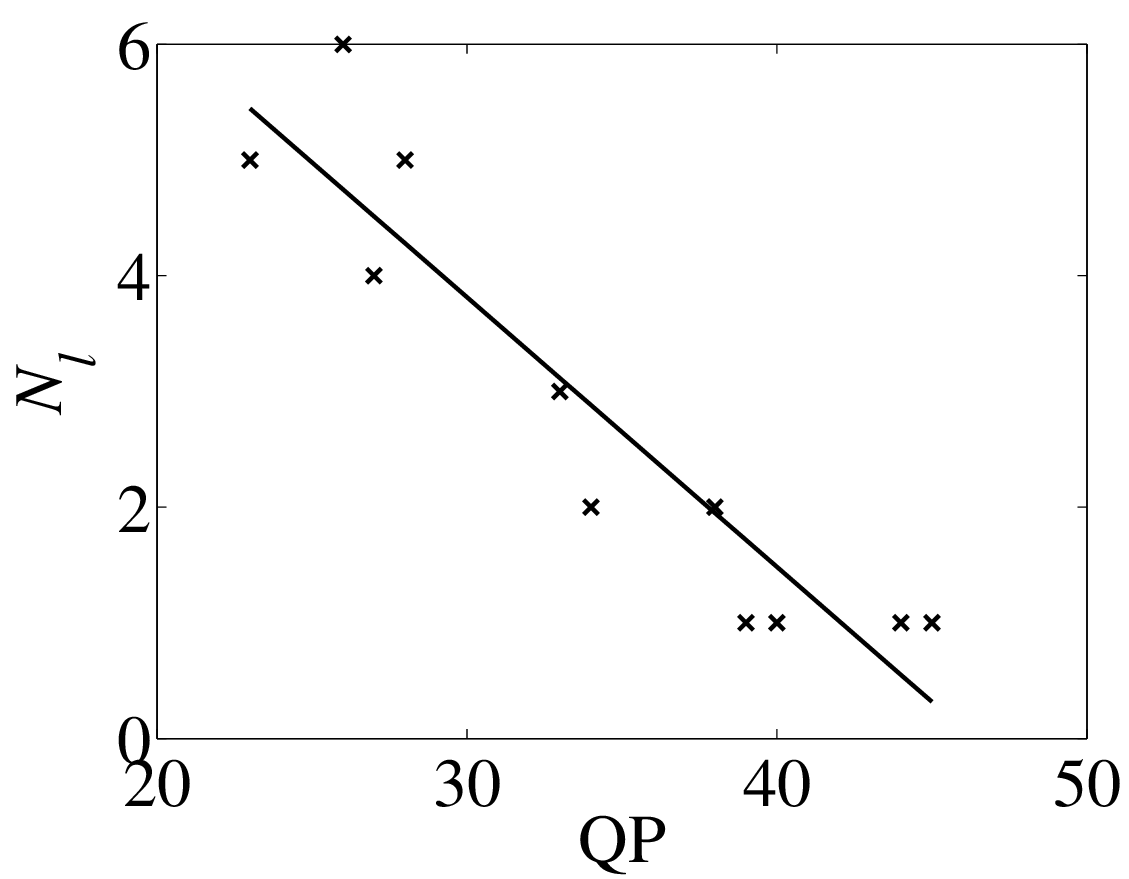}\\
			{\small (a)~1920x1080} & {\small (b)~1280x720} & {\small (b)~832x480} & {\small (b)~416x240}
		\end{tabular}
	\end{center}
    \vspace{-6mm}
	\caption{\label{fig:line_num} Relationship between the $N_l$ and QPs.}
	\vspace{-4mm}
\end{figure}

In the GALF filter coefficient training process~(\textit{S3}), the training of linear and nonlinear GALF~\cite{nonlinear-ALF} coefficients is conducted by the same scheme as shown in Fig.~\ref{Step3and4}~(a). In the first iteration, all the CTUs are allowed to be filtered, so the coefficients are calculated based on the auto-correlation matrix and cross-correlation vector of all the CTUs. Then, CTU-level on/off controls are decided using the derived filter coefficients based on a Rate-Distortion Optimization~(RDO) method. After this iteration, the filter coefficients are re-trained based on the filter-on CTUs. In the filter coefficient calculation process, $maxLumaFilters$~($maxLumaFilters = 25$) filters are firstly derived for the 25 luma classes. Then filter merging is conducted to derive the best number of Luma filter $N_l$. Though the distortion of the filtered image tends to be smaller with the increase of luma filter number, the luma filter coefficients usually demand more bits to be represented. To investigate the distribution of $N_l$, experiments are conducted on VTM-8.0 using the training set in Table~\ref{trainingSet}. As shown in Fig.~\ref{fig:line_num}, one can discern that the $N_l$ has a negative correlation with quantization parameters~(QP), which means that frames coded by lower QP tend to use fewer luma filters. The primary reason is that lower QP is usually corresponding to more flat areas and the smooth frame has a relatively simple structure. In view of this, we can reduce the $maxLumaFilters$ by predicting the most probable $N_l$ utilizing a resolution-dependent linear model. With the models, the $maxLumaFilters$ are set to be the predicted most probable $N_l$, which can adaptively simplify the coefficient calculation process.

\begin{figure}[t!]
	\begin{center}
		\noindent
		\begin{tabular}{cc}
			\epsfig{width=2.3in,file=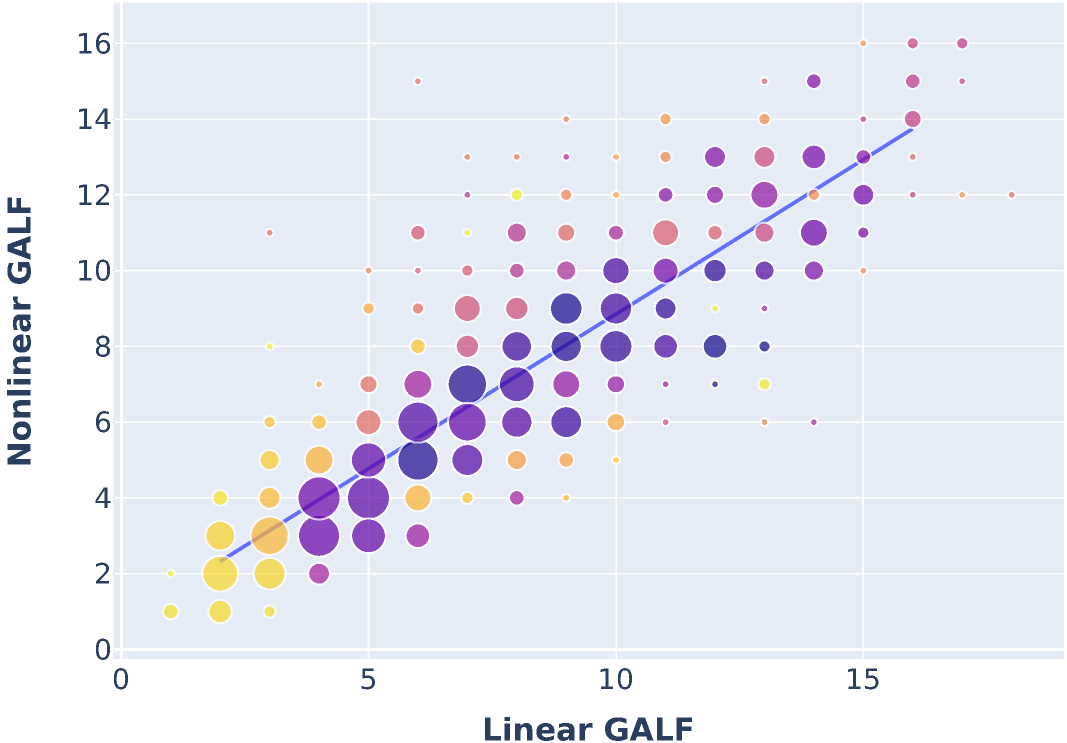} & 
			\epsfig{width=2.3in,file=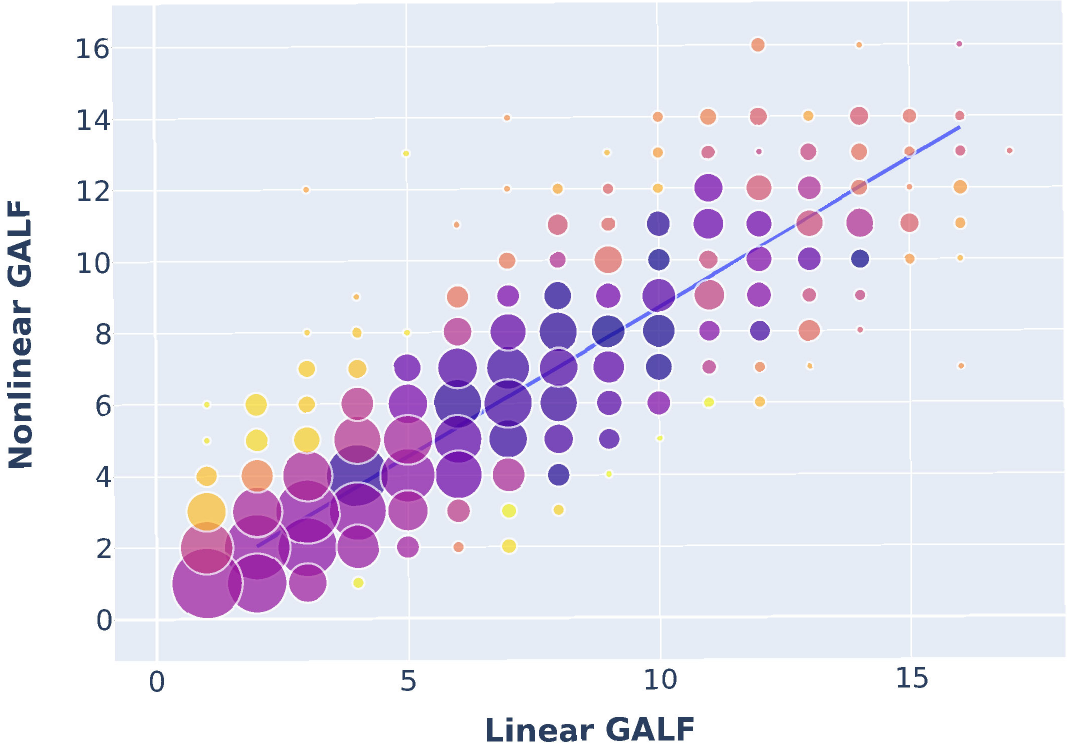} \\
			{\small (a)~I frame} & {\small (b)~B frame}
		\end{tabular}
	\end{center}
    \vspace{-6mm}
	\caption{Relationship between the $N_l$ of linear and nonlinear GALF.}\label{NonlinearToLinear}
	\vspace{-4mm}
\end{figure}

Apart from the relationship between $N_l$ and QP mentioned above, we also explore the correlation between $N_l$ of linear and nonlinear GALF, as shown in Fig.~\ref{NonlinearToLinear}. It can be seen that the $N_l$ of nonlinear GALF has a positive correlation with that of linear GALF, which means that we can refine the probable $N_l$ of nonlinear GALF with the existing $N_l$ derived by the linear GALF process. Hence, the nonlinear filtering process can be further simplified. 

For \textit{S4}, as shown in Fig.~\ref{Step3and4}~(b), three types of filter sets are available, including fixed filter sets trained off-line, temporal filter sets, and new filter set derived in \textit{S3}. In \textit{S4}, CTU-level filter is selected from fixed and temporal filter sets firstly in \textit{Stage1}. $T$ is the number of available temporal filter sets of the current frame. Then fixed, temporal and new filter sets are used for the CTU level decision in \textit{Stage2}. Let $fs_1$ and $fs_2$ to denote the filter sets selected in \textit{Stage1} and \textit{Stage2}, respectively. As the available filter sets in \textit{Stage1} and \textit{Stage2} are the same other than the new filter set, which is only used in \textit{Stage2}, we assume that the $fs_2$ are related to $fs_1$. To verify our conjecture, we conduct the test following the CTC. As shown in Table~\ref{step3_table}, the percentage is the filter set in $fs_2$ belonging to the $fs_1$. It is clear that all the selected filters other than the new filter in the \textit{Stage2} belong to $fs_1$. For B-frames, more than 97\% of the selected filters belong to the $fs_1$. So, the available filter sets in \textit{Stage2} can be reduced based on the $fs_1$ derived in \textit{Stage1}. 

\begin{table*}[t!]
	\centering
	\small
	\begin{center}
		\caption{Training and testing sequences.} \label{trainingSet}
		\begin{tabular}{c|c|c}
			\thickhline
			\hline
			Class & Training set & Testing set\\
			\hline
			\multicolumn{1}{c|} {\rule{0pt}{8pt}Class B}    & \textit{MarketPlace}, \textit{RitualDance}, \textit{Cactus} &  \textit{BasketballDrive}, \textit{BQTerrace}\\
			\multicolumn{1}{c|} {\rule{0pt}{8pt}Class C}    & \textit{BasketballDrill}, \textit{BQMall} & \textit{PartyScene}, \textit{RaceHorsesC}\\
			\multicolumn{1}{c|} {\rule{0pt}{8pt}Class D}    & \textit{BasketballPass}, \textit{BQSquare}& \textit{BlowingBubbles}, \textit{RaceHorses}\\
			\multicolumn{1}{c|} {\rule{0pt}{8pt}Class E}    & \textit{FourPeople}, \textit{Johnny}& \textit{KristenAndSara}\\
			\thickhline
		\end{tabular}
	\end{center}
	\vspace{-4mm}
\end{table*}

\begin{table*}[t!]
	\centering
	\small
	\begin{center}
		\caption{Percentage of the filter set in $fs_2$ belonging to the $fs_1$.} \label{step3_table}
		\begin{tabular}{c|c | c|c | c|c}
			\thickhline
			\hline
			& Class B& Class C&  Class D& Class E&  Average\\
			
			\hline
			\multicolumn{1}{c|} {\rule{0pt}{8pt}I frame} & 96.50\%  & 99.14\%   & 99.69\%       & 98.04\%      & 97.41\%        \\
			\multicolumn{1}{c|} {\rule{0pt}{8pt}B frame} & 100\%  & 100\%   & 100\%       & 100\%      & 100\%        \\
			\thickhline
		\end{tabular}
	\end{center}
	\vspace{-4mm}
\end{table*}

\begin{figure}[ht!]
	\begin{center}
		\noindent
		\includegraphics[width=5.5in]{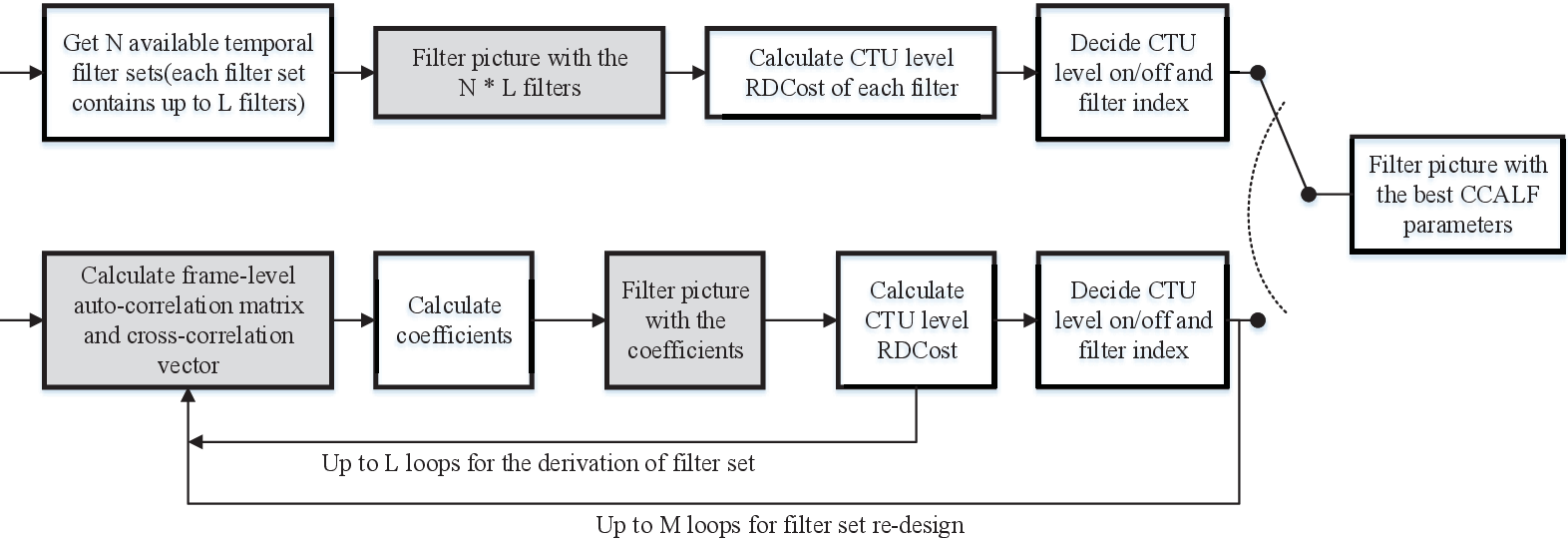}
        \vspace{-6mm}
		\caption{Multi-pass CCALF encoding flow in VTM-8.0.}\label{CCALF_workflow}
	\end{center}
	\vspace{-8mm}
\end{figure}

\SubSection{2.3 Analysis of CCALF Encoding Scheme}
The CCALF encoding process is shown in Fig.~\ref{CCALF_workflow}. There are up to $L$~($L=4$) filters in a frame and each CTU selects one filter from them. The filter parameter derivation process mainly includes two parts in the encoder, i.e., inheriting from the temporal parameter sets and training new parameters based on the current frame. When inheriting from the existing parameters, up to $N$~($N=8$) available filter sets can be used, and there are up to L filters in each filter set. Hence, the CCALF filtering process is conducted 32 times. In the new parameter training process, up to $M$~($M=15$) iterations are conducted to derive the filter set, and there are up to $L$ filters in the filter set. As the covariance calculation and filtering process are needed to access the buffer, there are up to 120 passes. So the total buffer access number is 152.

\Section{Optimized Adaptive Loop Filter Framework}
In this section, we propose the optimized ALF framework. First, we introduce the new encoding framework of the ALF module. Then the optimization of parameter training and CTU level filter selection process for the luma component in the GALF process is detailed. Finally, we propose a one-pass CCALF scheme to solve the problem caused by multiple picture buffer access.

\begin{figure}[t!]
 	\begin{center}
 		\noindent
 		\includegraphics[width=4.0in]{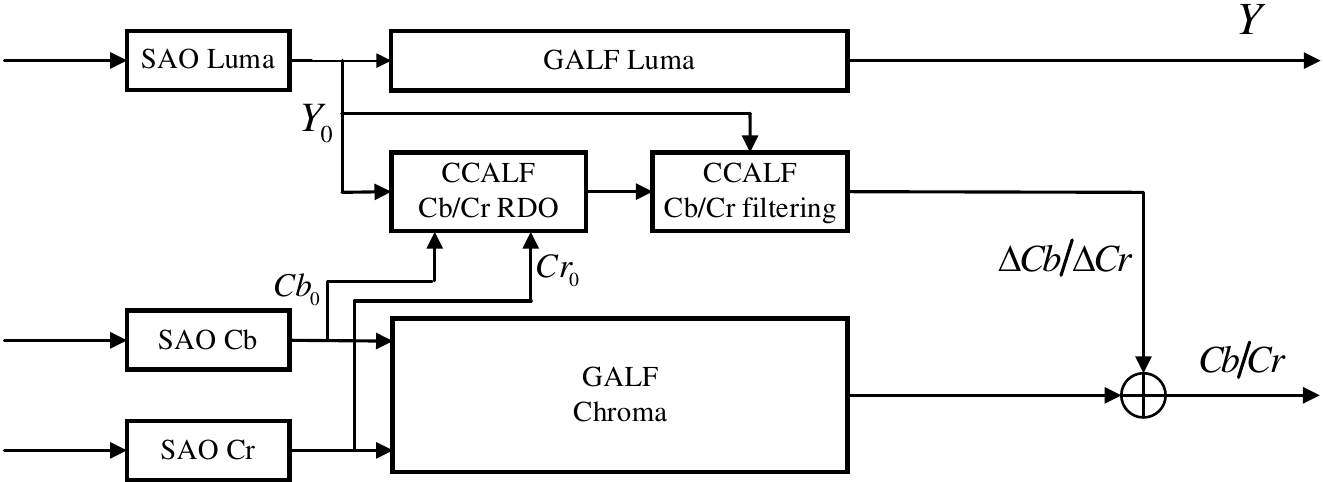}
        \vspace{-6mm}
 		\caption{Framework of the proposed ALF module in VTM-8.0.}\label{framework_pro}
 	\end{center}
 	\vspace{-8mm}
 \end{figure}

\SubSection{3.1 GALF and CCALF Encoder Parallel Design}
As the performance of CCALF is much higher than that of the GALF for chroma component, we assume that the GALF has less impact on chroma signal. So we propose to use chroma signal before GALF to replace the current reconstructed chroma signal in the CCALF parameter training process, as shown in Fig.~\ref{framework_pro}. With the proposed method, CCALF and GALF can be performed in parallel~\cite{Parallel}.

\SubSection{3.2 Optimized GALF for Luma Component}
The optimization of GALF mainly focuses on the adaptive parameter decision scheme of the two most time-consuming processes, i.e., coefficient training~(\textit{S3}) and CTU-level filter index decision process~(\textit{S4}). For \textit{S3}, we propose to estimate the parameter $maxLumaFilter$ using the off-line trained resolution-dependent linear model based on the observed relationship between $N_l$ and QP. Moreover, for the nonlinear GALF coefficient training process, the $maxLumaFilter$ is refined using the $N_l$ derived in the GALF process. These methods can efficiently reduce the unnecessary RDO process in \textit{S3}. For \textit{S4}, we propose to conduct the \textit{Stage2} process based on the results of \textit{Stage1}. Specifically, only the filters selected by the \textit{Stage1} and the new filter can be used as the candidates in the \textit{Stage2}. This scheme can remove unnecessary candidates in \textit{Stage2}. Consequently, the computational complexity is reduced.

\begin{figure}[t!]
	\begin{center}
		\noindent
		\includegraphics[width=5.5in]{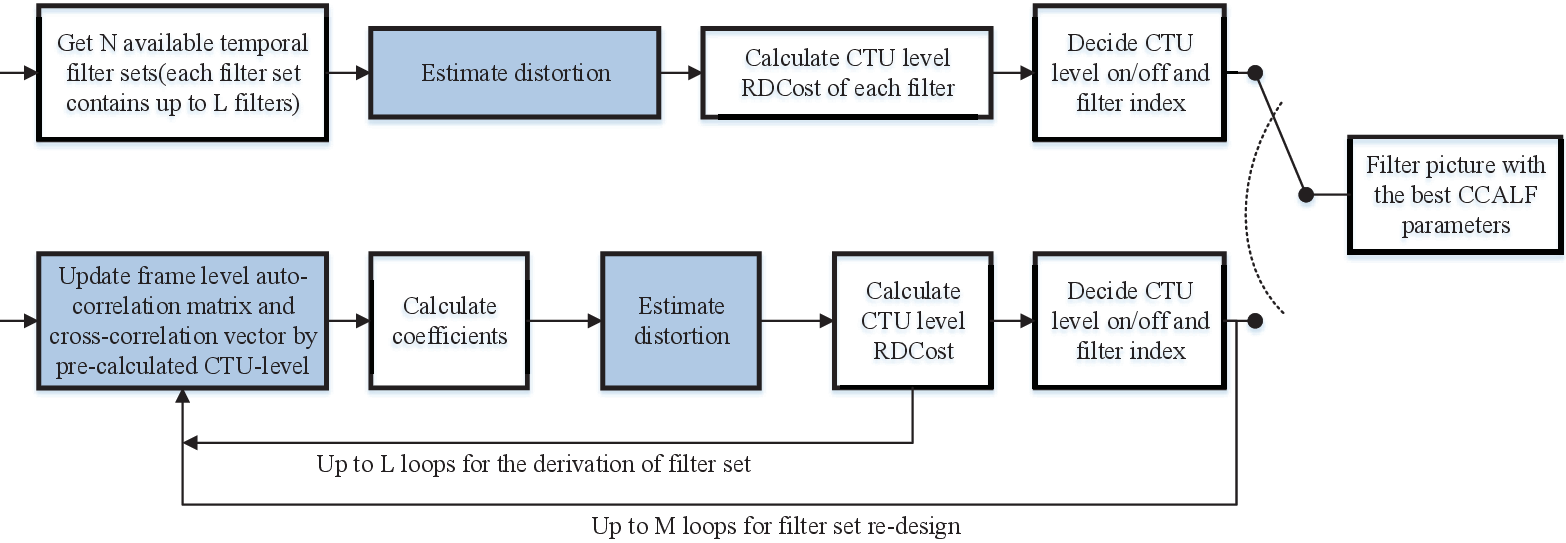}
        \vspace{-6mm}
		\caption{Proposed one-pass CCALF encoding flow in VTM-8.0.}\label{CCALF_workflow_pro}
	\end{center}
	\vspace{-10mm}
\end{figure}

\SubSection{3.3 One-pass CCALF Encoding Scheme}
As shown in Fig.~\ref{CCALF_workflow}, the auto-correlation matrix and cross-correlation vector calculation and filtering process in the parameter training process are the reason for the multi-pass CCALF scheme. Hence, how to redesign the two modules without accessing the image buffer is a crucial topic. So we propose the one-pass CCALF encoding scheme, which is shown in Fig.~\ref{CCALF_workflow_pro}.

For the auto-correlation matrix and cross-correlation vector calculation process, we propose to pre-calculate the auto-correlation matrix and cross-correlation vector of each CTU before the parameter training process. After that, the frame-level matrix and vector can be updated by that of the CTU level. The primary purpose of the filtering process is to calculate the distortion. So we propose a filtering distortion estimation method to derive the distortion without the filtered image. The mean-square error of CCALF can be calculated as follows~\cite{OnePass},
\begin{equation}\label{distortion1}
\varepsilon = E[(f[\textbf{k}] - s[\textbf{k}])^2] = E[(c[\textbf{k}] + \sum_{n=0}^{N-1} w_nl[\textbf{k}^{'}+\textbf{p}_n] - s[\textbf{k}])^2].
\end{equation}

After a series of substitutions, Equation~\eqref{distortion1} can be modified to 
\begin{equation}\label{distortion2}
\varepsilon = \dfrac{1}{\|K\|}\langle \textbf{w}, (\textbf{R}_{l,l}\textbf{w} - 2\textbf{R}_{l, s - c})\rangle + \dfrac{1}{\|K\|}\sum_{\textbf{k}\in K}(s[\textbf{k}] - c[\textbf{k}])^2, 
\end{equation}
where $\langle, \rangle$ is an inner product operation. $\|K\|$ is the number of pixels in $K$. $s[\textbf{k}]$ is the original sample of chroma component. $\textbf{R}_{l,l}$ denotes the auto-correlation matrix of $l[\textbf{k}^{'}]$. $\textbf{R}_{l,s-c}$ is the cross-correlation vector of $l[\textbf{k}^{'}]$ and the difference between $s[\textbf{k}]$ and $c[\textbf{k}]$. Hence, the distortion can be estimated by

\begin{equation}\label{distortion3}
\varepsilon = \langle \textbf{w}, (\textbf{R}_{l,l}\textbf{w} - 2\textbf{R}_{l, s - c})\rangle + \sum_{\textbf{k}\in K}(s[\textbf{k}] - c[\textbf{k}])^2.
\end{equation}

\begin{table*}[t!]
	\centering
	\small
	\begin{center}
		\caption{Estimation error of the proposed distortion estimation method.} \label{Estimation_Error}
		\begin{tabular}{c|c c|c c|c c}
			\thickhline
			\hline
			\multirow{2}{*}{\textbf{Class}}&
			\multicolumn{2}{c|}{\rule{0pt}{8pt} \textbf{AI}} & \multicolumn{2}{c}{\textbf{RA}}& \multicolumn{2}{c}{\textbf{LDB}}\\
			\cline{2-7} & 
			\textbf{U}& \textbf{V}&  \textbf{U}& \textbf{V}&  \textbf{U}& \textbf{V}\\
			
			\hline
			\multicolumn{1}{c|} {\rule{0pt}{8pt}Class A1} & 0.247\%  & 0.279\%   & 0.929\%       & 1.297\%      & -        & -\\
			\multicolumn{1}{c|} {\rule{0pt}{8pt}Class A2} & 0.113\%  & 0.089\%   & 0.678\%       & 0.653\%      & -        & -\\
			\multicolumn{1}{c|} {\rule{0pt}{8pt}Class B}  & 0.176\%  & 0.223\%   & 0.227\%  & 0.288\% & 0.213\%  & 0.282\%\\
			\multicolumn{1}{c|} {\rule{0pt}{8pt}Class C}  & 0.110\%  & 0.121\%   & 0.129\%  & 0.146\% & 0.114\%  & 0.125\%\\
			\multicolumn{1}{c|} {\rule{0pt}{8pt}Class E}  & 0.162\%  & 0.177\%   & -        & -       & 0.195\%  & 0.219\%\\
			\hline
			\multicolumn{1}{c|} {\rule{0pt}{8pt}\textbf{Average}} & 0.162\%  & 0.178\%   & 0.491\%  & 0.596\% & 0.174\%  & 0.209\%\\
			\thickhline
		\end{tabular}
	\end{center}
	\vspace{-8mm}
\end{table*}

To verify the effectiveness of this method, we calculate the estimation error by
\begin{equation}
error = abs(dist_E-dist_T)/ dist_T \times 100\%,
\end{equation}
where $dist_E$ is the estimated distortion and $dist_T$ is the distortion calculated by the filtered image and original image. The result is shown in Table~\ref{Estimation_Error}. It is clear that the estimation errors are very small and our estimation method is efficient.

\Section{Experimental Results}
In this section, we test the performance of the proposed algorithm in VTM-8.0. The videos in the CTC are utilized as test sequences in our experiment under three configurations, All Intra Main10~(AI), Random Access Main10~(RA) and Low-Delay B Main10~(LDB). The QP are set as {22, 27, 32, 37}. The first two seconds of these sequences are encoded for performance evaluation. Moreover, the coding performance is measured by Bjontegaard's method~\cite{BDrate} in terms of BD-rate.

In the experiment, the training set in Table~\ref{trainingSet} is used for the training of the relationship between the $N_l$ and QP. And the testing set in the table is used for the BD-rate calculation. As shown in Table~\ref{Performance1}, the proposed optimized ALF Module can achieve 22\%, 25\% and 24\% encoding time-savings of ALF module with negligible coding performance change for AI, RA and LDB configurations, respectively. The encoding time change compared to VTM-8.0 is minor. The reason is that the simulations are conducted by computer clusters, which cannot reflect the high external memory access latency caused by multi-pass CCALF. In practical ASIC design, the 152-pass encoding scheme will cause significantly longer latency and higher power consumption due to the frequent external memory access. Our proposed method is meaningful for the real-time encoder design and makes the high-performance ALF module more useful.

\begin{table*}[t!]
	\centering
	\small
	\begin{center}
		\caption{Experimental results of the proposed optimized ALF module~(Anchor: VTM-8.0).} \label{Performance1}
		\begin{tabular}{c|c c c|c c c|c c c}
			\thickhline
			\hline
			\multirow{2}{*}{\textbf{Class}}&
			\multicolumn{3}{c|}{\rule{0pt}{8pt} \textbf{AI}} & \multicolumn{3}{c}{\textbf{RA}}& \multicolumn{3}{c}{\textbf{LDB}}\\
			\cline{2-10} & 
			\textbf{Y}&  \textbf{U}& \textbf{V}&   \textbf{Y}&  \textbf{U}& \textbf{V}&   \textbf{Y}&  \textbf{U}& \textbf{V}\\
			
			\hline
			\multicolumn{1}{c|} {\rule{0pt}{8pt}Class B}    & 0.01\% & 0.19\%   & 0.09\%      &  0.06\%  & -0.51\%  & -0.30\% &  0.06\%  &  0.50\%  &  0.15\%\\
			\multicolumn{1}{c|} {\rule{0pt}{8pt}Class C}    & 0.00\% & 0.12\%   & 0.08\%      &  0.02\%  &  0.01\%  &  0.45\% & -0.06\%  &  0.48\%  & -0.42\%\\
			\multicolumn{1}{c|} {\rule{0pt}{8pt}Class D}    &-0.01\% & 0.13\%   & 0.08\%      & -0.02\%  &  0.05\%  & -0.01\% &  0.06\%  &  0.69\%  & 0.67\%\\
			\multicolumn{1}{c|} {\rule{0pt}{8pt}Class E}    & 0.00\% & 0.10\%   & 0.13\%      & -        & -        & -       & -0.22\%  &  0.71\%  & 0.00\%\\
			\hline
			\multicolumn{1}{c|} {\rule{0pt}{8pt}\textbf{Average}}     & 0.00\% & 0.14\%  &  0.09\%    & 0.02\%   & -0.15\%  & 0.05\% & -0.01\%  & 0.58\%  & 0.11\%\\
			\hline
			\multicolumn{1}{c|} {\rule{0pt}{8pt}$\Delta ET_{ALF}$} & \multicolumn{3}{c|}{\rule{0pt}{8pt} 78\%} & \multicolumn{3}{c|}{\rule{0pt}{8pt} 75\%} &
			\multicolumn{3}{c}{\rule{0pt}{8pt} 76\%}\\
			\hline
			\multicolumn{1}{c|} {\rule{0pt}{8pt}$\Delta ET_{all}$} & \multicolumn{3}{c|}{\rule{0pt}{8pt} 99\%} & \multicolumn{3}{c|}{\rule{0pt}{8pt} 98\%} &
			\multicolumn{3}{c}{\rule{0pt}{8pt} 98\%}\\
			\hline
			\multicolumn{1}{c|} {\rule{0pt}{8pt}$\Delta DT$} & \multicolumn{3}{c|}{\rule{0pt}{8pt}100\%} & \multicolumn{3}{c|}{\rule{0pt}{8pt} 100\%} &
			\multicolumn{3}{c}{\rule{0pt}{8pt} 100\%}\\
			\thickhline
		\end{tabular}
	\end{center}
\vspace{-8mm}
\end{table*}

\Section{Conclusion}
In this paper, an optimized ALF module, including the parallel design of GALF and CCALF, the adaptive parameter decision of GALF, and one-pass CCALF scheme is proposed. With the proposed method, the GALF and CCALF can be conducted in parallel in the encoder, and the CCALF encoding passes can be reduced from up to 152 to 1. Compared to VTM-8.0, the proposed framework achieves roughly 25\% time-savings of the ALF module with minor coding performance change under RA configuration. Our proposed method is meaningful for the real-time encoder design and makes the high-performance ALF module more practical.

\Section{References}
\footnotesize
\bibliographystyle{IEEEbib}
\bibliography{refs}

@inproceedings{non-local1,
  author={Meng, Xuewei and Jia, Chuanmin and Wang, Shanshe and Zheng, Xiaozhen and Ma, Siwei},
  booktitle={Picture Coding Symposium (PCS)}, 
  title={Optimized Non-local In-Loop Filter for Video Coding}, 
  year={2018},
  volume={},
  number={},
  pages={233-237},
  doi={10.1109/PCS.2018.8456299}}

@inproceedings{non-local2,
  title={Description of {SDR} video coding technology proposal by {DJI} and {P}eking {U}niversity},
  author={Wang, Zhao and Meng, Xuewei and Jia, Chuanmin and Cui, Jing and Wang, Suhong and Wang, Shanshe and Ma, Siwei},
  booktitle={document of Joint Video Experts Team JVET-J0011, 10th JVET Meeting},
  year={2018}
}

@inproceedings{non-local3,
  title={Non-local Structure-based Filter with integer operation},
  author={Meng, Xuewei and Jia, Chuanmin and Wang, Zhao and Ma, Siwei and Zheng, Xiaozhen},
  booktitle={Document of Joint Video Experts Team, JVET-J0071, 10th JVET Meeting},
  year={2018}
}

@inproceedings{non-local4,
  title={{CE}2: Non-local structure-based filter},
  author={Meng, Xuewei and Jia, Chunmin and Wang, Zhao and Ma, Siwei and Zheng, Xiaozhen},
  booktitle={Document of Joint Video Experts Team, JVET-K0160, 11th JVET Meeting},
  year={2018}
}

@inproceedings{CCALF22,
  title={{CE}5-related: High level syntax modifications for {CCALF} (combination of {JVET-Q0253} and {JVET-Q0520})},
  author={A. M. Kotra and S. Esenlik and B. Wang and H. Gao and others},
  booktitle={Document of Joint Video Experts Team, JVET-Q0782, 17th JVET Meeting},
  year={2019}
}

@inproceedings{r0322,
  title={{CCALF virtual boundary issue for 4:4:4 and 4:2:2 format}},
  author={Meng,Xuewei and Zheng, Xiaozhen and Wang, Shanshe and Ma, Siwei},
  booktitle={Joint Video Experts Team (JVET), document JVET-R0322},
  year={Oct. 2019}
}

@inproceedings{CCALF-1,
  title={{CE}5-related: On {CC-ALF} slice and picture header syntax},
  author={Meng, Xuewei and Zheng, Xiaozhen and Wang, Shanshe and Ma, Siwei},
  booktitle={Joint Video Experts Team (JVET) of ITU-T SG, JVET-Q0326, 17th JVET Meeting},
  year={2019}
}

@inproceedings{CCALF-2,
  title={{CE}5-related: High level syntax modifications for {CCALF} (combination of {JVET-Q0253} and {JVET-Q0520})},
  author={A. M. Kotra and S. Esenlik and B. Wang and H. Gao and others},
  booktitle={Joint Video Experts Team (JVET) of ITU-T SG, JVET-Q0782, 17th JVET Meeting},
  year={2019}
}

@inproceedings{NonlinearALF-1,
  title={Non-{CE5}: Unification of non-linear {ALF} luma/chroma clipping parameters},
  author={Meng, Xuewei and Zheng, Xiaozhen and Wang, Shanshe and Ma, Siwei},
  booktitle={Joint Video Experts Team (JVET) of ITU-T SG, JVET-O0437, 15th JVET Meeting},
  year={2019}
}

@inproceedings{NonlinearALF-2,
  title={Non-{CE5}: Modification of clipping value signalling for ALF},
  author={Meng, Xuewei and Zheng, Xiaozhen and Wang, Shanshe and Ma, Siwei},
  booktitle={Joint Video Experts Team (JVET) of ITU-T SG, JVET-O0430, 15th JVET Meeting},
  year={2019}
}

@article{HEVC,
	title={Overview of the {H}igh {E}fficiency {V}ideo {C}oding ({HEVC}) standard},
	author={Sullivan, Gary J and Ohm, Jens-Rainer and Han, Woo-Jin and Wiegand, Thomas},
	journal={IEEE Transactions on Circuits and Systems for Video Technology},
	volume={22},
	number={12},
	pages={1649--1668},
	year={2012}
}

@article{DF,
	title={Adaptive deblocking filter},
	author={List, Peter and Joch, Anthony and Lainema, Jani and Bjontegaard, Gisle and Karczewicz, Marta},
	journal={IEEE Transactions on Circuits and Systems for Video Technology},
	volume={13},
	number={7},
	pages={614--619},
	year={2003},
	publisher={IEEE}
}

@article{SAO,
	title={Sample {A}daptive {O}ffset in the {HEVC} standard},
	author={Fu, Chih-Ming and Alshina, Elena and Alshin, Alexander and others},
	journal={IEEE Transactions on Circuits and Systems for Video Technology},
	volume={22},
	number={12},
	pages={1755--1764},
	year={2012},
	publisher={IEEE}
}

@inproceedings{ALF-inloop,
	title={In-loop filter using block-based filter control for video coding},
	author={Watanabe, Takashi and Wada, Naofumi and Yasuda, Goki and Tanizawa, Akiyuki and others},
	booktitle={2009 16th IEEE International Conference on Image Processing (ICIP)},
	pages={1013--1016},
	year={2009},
	organization={IEEE}
}

@inproceedings{zheng2011directional,
	title={Directional adaptive loop filter for video coding},
	author={Zheng, Yunfei and Yin, Peng and Xu, Qian and Sole, Joel and Lu, Xiaoan},
	booktitle={2011 18th IEEE International Conference on Image Processing},
	pages={3501--3504},
	year={2011},
	organization={IEEE}
}

@article{nonlinear-ALF,
	title={{CE}5: {R}esults of tests {CE}5-3.1 to {CE}5-3.4 on {N}on-Linear {A}daptive {L}oop {F}ilter},
	author={Taquet, Jonathan and Onno, Patrice and Gisquet, Christophe and Laroche, Guillaume},
	journal={JVET-N0242, 14th JVET Meeting, Geneva, CH},
	year={2019}
}

@article{R0013,
	title={{JVET AHG} report: {T}ool reporting procedure({AHG}13)},
	author={Chien, Wei-Jung and Boyce, Jill and others },
	journal={JVET-R0013, 18th JVET Meeting, teleconference},
	year={2020}
}

@article{CC-ALF,
	title={{CC-ALF}: {I}ntegrated {T}ext for the {C}ross {C}omponent {A}daptive {L}oop {F}ilter},
	author={Misra, Kiran and Bossen, Frank and Segall, Andrew and others},
	journal={JVET-Q0795, 17th JVET Meeting, Brussels, BE},
	year={2020}
}

@article{GALF,
	title={{CE}2.4.1.4: {R}educed filter shape size for {ALF}},
	author={Karczewicz, Marta and Shlyakhov, Nikolay and Hu, Nan and Seregin, Vadim and Chien, Wei-Jung},
	journal={JVET-K0371, 11th JVET Meeting, Ljubljana, SI},
	year={2018}
}

@article{VTM8.0,
	title={Algorithm description for {V}ersatile {V}ideo {C}oding and {T}est {M}odel 8 ({VTM} 8)},
	author={Chen, Jianle and Ye, Yan and Kim, Seung Hwan},
	journal={JVET document, JVET-Q2002},
	year={2020}
}

@article{OnePass,
	title={{AHG} 10: {O}ne-pass {CCALF}},
	author={Meng, Xuewei and Zheng, Xiaozhen and Wang, Shanshe and Ma, Siwei},
	journal={JVET document, JVET-R0327},
	year={2020}
}

@article{R0225,
	title={{AHG}9: On {ALF/CC-ALF} high level syntax},
	author={Meng, Xuewei and Zheng, Xiaozhen and Wang, Shanshe and Ma, Siwei},
	journal={JVET document, JVET-R0225},
	year={2020}
}

@article{Parallel,
	title={{AHG} 10: {ALF} and {CCALF} encoder parallel design},
	author={Meng, Xuewei and Zheng, Xiaozhen and Wang, Shanshe and Ma, Siwei},
	journal={JVET document, JVET-R0328},
	year={2020}
}

@article{CFP,
	title={Joint Call for Proposals on Video Compression with Capability beyond {HEVC}},
	author={Segall, Andrew and Baroncini, Vittorio and Boyce, Jill and Chen, Jianle and Suzuki, Teruhiko},
	journal={JVET document, JVET-H1002},
	year={2017}
}

@article{CTC,
	title={{JVET} common test conditions and software reference configurations for {SDR} video},
	author={Bossen, Frank and Boyce, Jill and Suehring, Karsten and Li, Xiang and Seregin, Vadim },
	journal={JVET document, JVET-N1010},
	year={2019}
}

@article{BDrate,
	title={Calculation of average {PSNR} differences between {RD}-curves},
	author={Bjontegarrd, Gisle},
	journal={VCEG-M33, 13th VCEG Meeting, Austin, TX, USA},
	year={2001}
}

@Misc{fraunhoferhhi,
	howpublished = {},
	note = {\url{https://vcgit.hhi.fraunhofer.de/jvet/VVCSoftware\_VTM/tags/VTM\-4.0} Accessed October 28, 2020},
	title = {fraunhoferhhi/vvenc},
	author = {Fraunhofer HHI}
}

@article{tsai2013adaptive,
	title={Adaptive loop filtering for video coding},
	author={Tsai, Chia-Yang and Chen, Ching-Yeh and Yamakage, Tomoo and Chong, In Suk and Huang, Yu-Wen and others},
	journal={IEEE Journal of Selected Topics in Signal Processing},
	volume={7},
	number={6},
	pages={934--945},
	year={2013},
	publisher={IEEE}
}

\end{document}